\documentclass[acmtog,authorversion]{acmart}

\usepackage{booktabs}
\usepackage{graphicx}
\usepackage{enumitem}
\usepackage{soul}
\usepackage{dsfont}
\usepackage{gensymb}
\usepackage{microtype}
\usepackage{pifont}
\usepackage{threeparttable}
\usepackage{collect}
\usepackage{xspace}
\usepackage{xfrac}
\usepackage[capitalise]{cleveref}
\usepackage{natbib}
\usepackage{array}
\usepackage{multirow}
\usepackage{wrapfig}
\usepackage{url}
\usepackage{icomma}
\usepackage{mathdots}

\usepackage{pifont}

\newcommand{\xmark}{\scalebox{0.85}{\ding{53}}}

\DeclareGraphicsExtensions{.png,.jpg,.pdf,.ai,.psd}
\newcommand{\eg}{e.g.,\ }
\newcommand{\ie}{i.e.,\ }

\creflabelformat{equation}{#2\textup{#1}#3}
\newcommand{\refFig}[1]{Fig.~\ref{fig:#1}}

\newcommand{\refTab}[1]{Tab.~\ref{tab:#1}}

\newcommand{\refSec}[1]{Sec.~\ref{sec:#1}}

\newcommand{\refEq}[1]{Eq.~\ref{eq:#1}}

\DeclareMathOperator*{\argmin}{arg\,min}

\definecollection{mymaths}
\newcommand{\mymath}[2]{
    \newcommand{#1}{\TextOrMath{$#2$\xspace}{#2}}
    \begin{collect}{mymaths}{}{}{}{}
    #1
    \end{collect}
}
\mymath{\signal}{f}
\mymath{\field}{F}
\mymath{\indim}{{d_i}}
\mymath{\outdim}{{d_o}}
\mymath{\sensordim}{{d_s}}
\mymath{\coord}{\mathbf{x}}
\mymath{\sensorcoord}{{\mathbf{x}_s}}
\mymath{\offset}{\boldsymbol\tau}
\mymath{\covariance}{\Sigma}
\mymath{\pseudocovariance}{{\hat{\Sigma}}}
\mymath{\scalespacesignal}{{\signal_\covariance}}
\mymath{\lipschitzbound}{c}
\mymath{\layercount}{l}
\mymath{\weightmatrix}{W}
\mymath{\biasvector}{\mathbf{b}}
\mymath{\positionalencoding}{\gamma}
\mymath{\modulatedpositionalencoding}{\positionalencoding}
\mymath{\fourierfeature}{\mathbf{a}}
\mymath{\fourierfeaturecount}{m}
\mymath{\fourierfeatureweight}{\lambda}
\mymath{\trainableweights}{\theta}
\mymath{\mlp}{\Psi_\trainableweights}
\mymath{\boundedmlp}{{\overline\mlp}}
\mymath{\eigenvectormatrix}{Q}
\mymath{\eigenvaluematrix}{\Lambda}
\mymath{\forwardmap}{\Phi}
\mymath{\leftsingularmatrix}{U}
\mymath{\singularmatrix}{S}
\mymath{\rightsingularmatrix}{V}
\mymath{\skewsymmetricmatrix}{A}
\mymath{\covariancetransformation}{h}
\mymath{\variance}{\sigma^2}
\mymath{\pseudovariance}{\hat{\sigma}^2}
\mymath{\coordsamples}{{n_{\coord}}}
\mymath{\variancesamples}{{n_{\variance}}}
\mymath{\pseudovariancesamples}{{n_{\pseudovariance}}}
\mymath{\mcsamples}{N}
\mymath{\correctionfactor}{\mu}
\mymath{\fourierwarpingvariance}{{\sigma^2_\fourierfeature}}

\setcopyright{rightsretained}
\acmJournal{TOG}
\acmYear{2024}
\acmVolume{43}
\acmNumber{4}
\acmArticle{134}
\acmMonth{7}
\acmDOI{10.1145/3658163}

\begin{document}

\title{Neural Gaussian Scale-Space Fields}

\author{Felix Mujkanovic}
\affiliation{
  \institution{Max-Planck-Institut für Informatik}
  \country{Germany}
}
\email{felix.mujkanovic@mpi-inf.mpg.de}
\orcid{0009-0009-9122-4408}

\author{Ntumba Elie Nsampi}
\affiliation{
  \institution{Max-Planck-Institut für Informatik}
  \country{Germany}
}
\email{nnsampi@mpi-inf.mpg.de}
\orcid{0000-0002-3769-9850}

\author{Christian Theobalt}
\affiliation{
  \institution{Max-Planck-Institut für Informatik}
  \country{Germany}
}
\email{theobalt@mpi-inf.mpg.de}
\orcid{0000-0001-6104-6625}

\author{Hans-Peter Seidel}
\affiliation{
  \institution{Max-Planck-Institut für Informatik}
  \country{Germany}
}
\email{hpseidel@mpi-sb.mpg.de}
\orcid{0000-0002-1343-8613}

\author{Thomas Leimkühler}
\affiliation{
  \institution{Max-Planck-Institut für Informatik}
  \country{Germany}
}
\email{thomas.leimkuehler@mpi-inf.mpg.de}
\orcid{0009-0006-7784-7957}

\begin{teaserfigure}
    \includegraphics[width=\textwidth]{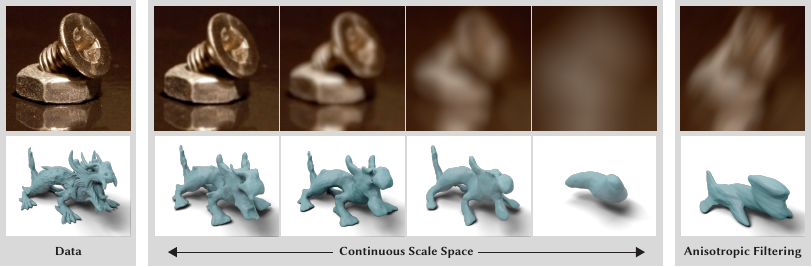}
    \caption{We learn neural fields that capture continuous, anisotropic Gaussian scale spaces. Given a training signal, such as an image or geometry (left), we learn a neural field representation that allows continuous Gaussian smoothing (center). Crucially, this representation is learned self-supervised, \ie without ever filtering the training signal. Our scale spaces are continuous in all parameters, including arbitrary covariance matrices that allow anisotropic filtering (right).}
    \label{fig:teaser}
    \Description{}
\end{teaserfigure}

\begin{abstract}
Gaussian scale spaces are a cornerstone of signal representation and processing, with applications in filtering, multiscale analysis, anti-aliasing, and many more. However, obtaining such a scale space is costly and cumbersome, in particular for continuous representations such as neural fields. We present an efficient and lightweight method to learn the fully continuous, anisotropic Gaussian scale space of an arbitrary signal. Based on Fourier feature modulation and Lipschitz bounding, our approach is trained self-supervised, \ie training does not require any manual filtering. Our neural Gaussian scale-space fields faithfully capture multiscale representations across a broad range of modalities, and support a diverse set of applications. These include images, geometry, light-stage data, texture anti-aliasing, and multiscale optimization.
\end{abstract}



\maketitle

\vspace{10mm}
\section{Introduction}
\label{sec:introduction}

Continuous neural representations, so-called neural fields, are becoming ubiquitous in visual-computing research and applications \cite{xie2022neural,tewari2022advances}. At their core, they map coordinates to signal values using a neural network. The generality, compactness, and malleability of this continuous data structure make them a popular choice for representing a broad variety of modalities. For example, neural fields have been used to represent geometry~\cite{park2019deepsdf}, images~\cite{stanley2007compositional}, radiance fields~\cite{mildenhall2020nerf}, flow~\cite{park2021nerfies}, reflectance~\cite{gargan1998approximating}, and much more.

In its basic form, a trained neural field allows querying the \emph{original} signal value for a given coordinate. Oftentimes, however, this functionality is not sufficient: Many important use cases require \emph{low-pass filtered versions} of the signal, arising from a custom band-limiting kernel that might be spatially-varying and even anisotropic. For example, a downstream application might require querying the signal at different scales \cite{marr1980theory,starck1998image}, or a subsequent discretization stage demands careful pre-filtering for anti-aliasing \cite{greene1986creating,antoniou2006digital}.

Scale-space theory \cite{iijima1959basic,lindeberg2013scale,witkin1987scale,koenderink1984structure} provides a principled framework for tackling this problem. A linear scale-space representation is obtained by creating a family of progressively Gaussian-smoothed versions of a signal. This effectively leads to a progressive suppression of fine-scale structures. Once this representation is created, filtering boils down to merely querying the scale space at the required location. In this work, we set out to develop a lightweight and efficient method to learn a neural field that captures a fully continuous, anisotropic Gaussian scale space in a self-supervised manner.

Obtaining such a representation is non-trivial, as na\"{\i}vely executing large-scale (anisotropic) Gaussian convolutions is highly inefficient. The task is even more challenging in neural fields, as they typically only allow point-wise function evaluations. Gaussian-weighted aggregation can be achieved using Monte Carlo estimation, but this requires trading high computational cost against high variance. Recently, convolutions have been rather efficiently executed in neural fields using differentiation strategies~\cite{xu2022signal,Nsampi2023NeuralFC}. Yet, these solutions only support either fixed small-scale kernels and require costly repeated automatic differentiation~\cite{xu2022signal}, or are limited to axis-aligned kernels and demand multiple forward passes per filter location~\cite{Nsampi2023NeuralFC}. Neural fields can directly learn a continuous, isotropic scale space~\cite{barron2021mipnerf,barron2022mip}, but typically require dense supervision across scales. Specialized network architectures allow to learn an explicit decomposition of the signal into different frequency bands~\cite{lindell2022bacon,fathony2020multiplicative,shekarforoush2022residual,saragadam2022miner}, but only support a coarse, discrete set of isotropic scales. 

In this work, we present a novel approach for a neural field to learn a complete anisotropic Gaussian scale space that is applicable to arbitrary signals and modalities~(\refFig{teaser}). Different from previous works, our scale spaces are \emph{fully continuous in all parameters}, \ie both in signal coordinates and in arbitrary Gaussian covariance matrices. This allows fine-grained, spatially-varying (pre-)filtering using only a single forward pass. Crucially, training is self-supervised, \ie we do not require supervision from filtered versions of the training signal, facilitating lightweight training and, consequently, a broad applicability of our method.

We observe that a positional encoding in the form of Fourier features~\cite{rahimi2007random,tancik2020fourfeat,hertz2021sape} provides a convenient means to modulate frequency content. However, the key ingredient to allow high-quality low-pass filtering of a signal is to pair this encoding with a Lipschitz-bounded multi-layer perceptron (MLP)~\cite{miyato2018spectral,gouk2021regularisation,szegedy2013intriguing}. We show that this regularization translates the dampening of encoding frequencies into a Gaussian smoothing of the signal. Training such an MLP with anisotropically modulated Fourier features on the \emph{original} training signal forces the network to learn a Gaussian scale space, without requiring any manual filtering. After training, a calibration stage maps modulation parameters to Gaussian variance, based on ultra-lightweight, one-time Monte Carlo estimates of Gaussian convolutions. At inference, filtered versions of the learned signal can be synthesized in a single forward pass using arbitrary, continuous Gaussian covariance matrices.

We evaluate the accuracy and applicability of our approach on a broad variety of tasks and modalities. This includes anisotropic smoothing of images and geometry; (pre-)filtering of textures and light-stage data; spatially varying filtering; and multiscale optimization. We further analyze the interplay between Fourier features and Lipschitz-bounded MLPs to elucidate the combined effect of the central ingredients of our approach.

In summary, our contributions are:
\begin{itemize}
    \item A novel approach for learning a fully continuous, anisotropic Gaussian scale space in a general-purpose neural field representation.
    \item An effective and efficient training methodology to achieve this goal self-supervised, \ie without the requirement to filter the training data.
    \item The application and careful evaluation of our method on a spectrum of relevant modalities and tasks.
\end{itemize}

\section{Related Work}
\label{sec:relatedwork}

Here, we review related work on classical and neural multiscale representations (\refSec{related_multiscale_representations}), the use of Fourier features in neural fields (\refSec{related_fourier}), and Lipschitz bounds in deep learning (\refSec{related_lipschitz}). For a comprehensive overview of neural fields in visual computing, we refer to recent surveys~\cite{xie2022neural,tewari2022advances}.

\subsection{Multiscale Signal Representations}
\label{sec:related_multiscale_representations}

Representing a signal at multiple scales has a long history in signal processing and visual computing, with the concept of scale spaces \cite{iijima1959basic,witkin1987scale,koenderink1984structure,lindeberg2013scale} at the center of attention. Many different scale spaces can be constructed from a signal~\cite{weickert1998anisotropic,florack1995nonlinear,dorst1994morphological} using a rigorous axiomatic foundation~\cite{lindeberg1997axiomatic}. Of special interest is the linear form, \ie the convolution of the signal with a family of Gaussian kernels, as it exhibits a number of useful and well-studied properties, such as predictable behavior after differentiation~\cite{babaud1986uniqueness}.

Scale spaces are typically constructed using various flavors of discretization. The convolution of a discrete signal with a discretized Gaussian kernel can be executed using cubature, but comes at high computational cost, especially in higher dimensions. Acceleration strategies involve the discrete Fourier transform~\cite{brigham1988fast} or exploiting the separability of Gaussians~\cite{geusebroek2003fast}. In low dimensions, pyramidal structures (MIP mapping)~\cite{burt1981fast,williams1983pyramidal} are even more efficient. Here, the isotropic scale parameter, \ie the pyramid level, is also discrete. Anisotropic filtering using pyramids (RIP mapping) exists~\cite{simoncelli1995steerable}, but comes at the cost of an additional coarse discretization of filter orientation, which can be hidden using carefully designed steerable filters~\cite{freeman1991design}. A different line of work has successfully explored signal representations using a discrete set of multiscale basis functions~\cite{daubechies1988orthonormal,mallat1989theory,guo2006sparse}. In contrast to all these works, our approach is fully continuous in all dimensions.

Continuous representations impose significant challenges for multiscale techniques, and a dominant strategy for filtering is stochastic multi-sampling~\cite{hermosilla2018monte,shocher2020discrete,wang2018deep,barron2023zipnerf,ma2022deblur}. Such Monte Carlo approaches require a high number of samples to avoid objectionable noise. Sample count can be significantly reduced by relying on differentiation and integration properties of convolutions~\cite{Nsampi2023NeuralFC}. However, learning the required integral representation is costly, high-quality Gaussian filtering still requires a substantial number of network evaluations, and general anisotropic filtering requires pre-computing many kernel shapes. Our neural fields are easy to train and allow arbitrary anisotropic Gaussian filtering using just a single forward pass. A different strategy relies on approximating continuous filtering using a learned linear combination of derivatives of the signal obtained via automatic differentiation~\cite{xu2022signal}. Different from our solution, this approach only supports small filter kernels.

Pre-filtering for anti-aliasing in continuous neural representations has recently received a lot of attention~\cite{barron2021mipnerf,barron2022mip,hu2023Tri-MipRF,nam2023mip}. Similar to our approach, these solutions employ carefully crafted inductive biases that help learn a multiscale representation. However, they rely on supervision across scales, such as images of scene objects captured at different distances. In contrast, our method allows to learn a full scale space from a single-scale supervision signal, thereby significantly extending its applicability to a broad range of signals and modalities.

Strong architectural inductive biases allow training neural networks with intermediate activations that represent progressively band-limited versions of the learned signal~\cite{fathony2020multiplicative,lindell2022bacon,shekarforoush2022residual}, with applications in coarse-to-fine learning~\cite{karras2017progressive,xiangli2022bungeenerf}. This leads to a discretization of scales and typically only allows isotropic filtering with a $\mathrm{sinc}$-kernel. Extending this scheme to the anisotropic case is possible~\cite{yang2022polynomial}, but, similar to RIP mapping, it requires an additional coarse discretization of filter orientation, limiting this approach to low dimensions. The discretization of scales can be combined with neuroexplicit architectures, \eg via spatial discretization or subdivision~\cite{saragadam2022miner,takikawa2022variable}, and many corresponding domain-specific solutions exist~\cite{chen2021learning,xu2021ultrasr,paz2022multiresolution,kuznetsov2021neumip,GFLTTB:2022:MIPNet,takikawa2021nglod,chen2023neural,zhuang2023anti}. Yet, none of them allows fully continuous, arbitrary, anisotropic Gaussian filtering.

\subsection{Fourier Features in Neural Fields}
\label{sec:related_fourier}

With early applications in time series analysis and representation learning~\cite{kazemi2019time2vec,vaswani2017attention,xu2019self}, Fourier features~\cite{rahimi2007random} are now a popular tool for learning neural-field representations of signals~\cite{mildenhall2020nerf}. Also referred to as positional encoding, their ability to map coordinates to latent features of different frequencies is an effective remedy for the spectral bias of neural networks~\cite{rahaman2019spectral}. \citet{tancik2020fourfeat} have analyzed the properties of such an encoding for neural fields using the neural tangent kernel~\cite{jacot2018neural}, and propose the use of normally distributed frequency vectors. Many applications rely on carefully dampened Fourier features to increase training stability~\cite{hertz2021sape,park2021nerfies,lin2021barf,yang2023freenerf}, or to obtain a multiscale representation given a multiscale supervision signal~\cite{barron2021mipnerf, barron2022mip}. Further, the analytical structure of Fourier features has been exploited for alias-free image synthesis~\cite{karras2021alias}. We employ dampened Fourier features with carefully chosen frequency vectors as well, and combine this encoding with a Lipschitz-bounded network to obtain a Gaussian scale space.

Fourier features have been explored in different architectural variants. Examples of this scheme include periodic activation functions~\cite{sitzmann2020implicit,mehta2021modulated}, Wavelet-style spatio-spectral encodings~\cite{wu2023neural}, or the modulation of unstructured representations based on radial basis functions~\cite{chen2023neurbf}. Different from our approach, the goal of these works is to improve single-scale reconstruction quality of complex signals.

\subsection{Lipschitz Networks and Matrix Parameterizations}
\label{sec:related_lipschitz}

Neural networks with guaranteed Lipschitz bounds have numerous applications, such as robustness~\cite{cisse2017parseval,hein2017formal}, smooth interpolation~\cite{10.1145/3528233.3530713}, and generative modeling~\cite{arjovsky2017wasserstein}. Technically, a desired Lipschitz constant can be enforced on the level of individual weight matrices. Corresponding methods can be divided into two classes.

The first class of methods relies on variants of projected gradient descent, where weight matrices are projected towards the closest feasible solution for each optimization step~\cite{szegedy2013intriguing}. This can be done using spectral normalization~\cite{yoshida2017spectral,miyato2018spectral,gouk2021regularisation,behrmann2019invertible,yang2021geometry}, or using more sophisticated projections~\cite{cisse2017parseval} relying on orthonormalization~\cite{bjorck1971iterative}. The second class of methods reparameterize the weight matrices such that an unconstrained optimization can be applied~\cite{anil2019sorting,10.1145/3528233.3530713}. Our method relies on this approach, as we observe that it leads to controllable training dynamics, but the choice of matrix parameterization is crucial for numerical stability.

The singular value decomposition is a convenient tool in this context~\cite{zhang2018stabilizing,mathiasen2020if}. As it requires a parameterization of orthogonal matrices, ad-hoc parameterizations for special orthogonal matrices have been considered~\cite{helfrich2018orthogonal,huang2018orthogonal,jing2017tunable,arjovsky2016unitary}. More general solutions rely on Householder reflections~\cite{mhammedi2017efficient,zhang2018stabilizing,mathiasen2020if}, but they exhibit unfavourable properties when used within an optimization loop. We rely on matrix exponentials~\cite{lezcano2019cheap,hyland2017learning}, which have been shown to outperform other approaches~\cite{golinski2019improving}. Yet, to the best of our knowledge, we are the first to use this set of techniques in the context of Lipschitz-bounded neural networks.

\section{Preliminaries}
\label{sec:preliminaries}

Here, we introduce concepts relevant to our approach. We first establish our notation for signals and fields, before reviewing technical background on Gaussian scale spaces, Fourier features, and Lipschitz continuity.

\paragraph{Signals and fields}
We are concerned with arbitrary continuous signals $\signal \in \mathds{R}^{\indim} \rightarrow \mathds{R}^{\outdim}$, where \indim and \outdim are typically rather small. This generic formulation captures many modalities in visual computing, \eg RGB images using $\indim=2$, $\outdim=3$, or signed distance functions (SDFs) for geometry using $\indim=3$, $\outdim=1$. Ultimately, we are interested in fitting a neural field $\field \in \mathds{R}^{\indim} \rightarrow \mathds{R}^{\outdim}$ 
to (versions of) \signal. We write continuous coordinates of signals and fields as $\coord \in \mathds{R}^{\indim}$.

\begin{figure}
    \includegraphics[width=0.99\linewidth]{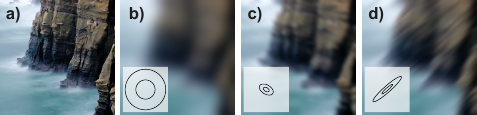}
    \caption{An original 2D signal \signal (\emph{a}) alongside samples from its Gaussian scale space \scalespacesignal (\emph{b}-\emph{d}). In \emph{b}), isotropic smoothing is applied, while \emph{c}) and \emph{d}) demonstrate examples of anisotropic filtering. Insets show isolines of the corresponding Gaussian kernels. We consider scale spaces that are continuous both in signal coordinates and in full Gaussian covariance.}
    \label{fig:scale_space_example}
    \Description{}
\end{figure}

\paragraph{Gaussian scale space}
The linear (Gaussian) scale space~\cite{iijima1959basic,lindeberg2013scale,witkin1987scale,koenderink1984structure} of \signal is defined as the continuous convolution of \signal with a Gaussian kernel with positive definite covariance matrix $\covariance \in \mathds{R}^{\indim \times \indim}$:
\begin{equation}
    \label{eq:scale_space_signal}
    \scalespacesignal(\coord)
    \coloneqq
    \frac
        {1}
        {\sqrt{(2 \pi)^\indim \det(\covariance)}}
    \int_{\mathds{R}^\indim}
    \signal(\coord - \offset)
    \exp
    \left(
        -\frac{1}{2}
        \offset^T
        \covariance^{-1}
        \offset
    \right)
    \mathrm{d} \offset.
\end{equation}
In such an augmented signal representation, continuously varying \covariance gives rise to differently smoothed versions of the signal~(\refFig{scale_space_example}). A very practical benefit of this structure is that it provides precise control over the frequency content of a signal. Yet, obtaining a scale-space representation is costly, as it requires executing a continuous \indim-dimensional integral for each combination of \coord and $\covariance$ -- an operation for which, in general, no closed-form solution exists.

\paragraph{Fourier features}
Neural networks exhibit an intrinsic spectral bias towards ``simple'' solutions~\cite{rahaman2019spectral}, which makes it challenging for basic fully-connected architectures to learn high-frequency content. An established remedy is to first featurize the input coordinate \coord using a fixed mapping $\positionalencoding \in \mathds{R}^{\indim} \rightarrow \mathds{R}^{2\fourierfeaturecount}$, based on sinusoids of \fourierfeaturecount different frequencies~\cite{mildenhall2020nerf,tancik2020fourfeat,rahimi2007random}:
\begin{equation}
    \label{eq:positional_encoding}
    \positionalencoding(\coord)
    =
    \begin{pmatrix}
        \fourierfeatureweight_1 \cos\left( 2 \pi \fourierfeature_1^T \coord \right)
        \\
        \fourierfeatureweight_1 \sin\left( 2 \pi \fourierfeature_1^T \coord \right)
        \\
        \vdots
        \\
        \fourierfeatureweight_\fourierfeaturecount \cos\left( 2 \pi \fourierfeature_\fourierfeaturecount^T \coord \right)
        \\
        \fourierfeatureweight_\fourierfeaturecount \sin\left( 2 \pi \fourierfeature_\fourierfeaturecount^T \coord \right)
    \end{pmatrix}.
\end{equation}
In this positional encoding, $\fourierfeature_i \in \mathds{R}^\indim$ are frequency vectors and $\fourierfeatureweight_i \in \mathds{R}$ are weights of the corresponding Fourier feature dimensions. Small offsets in \coord lead to rapid changes of $\positionalencoding(\coord)$ for high frequencies $\fourierfeature_i$. Therefore, feeding $\positionalencoding(\coord)$ instead of the raw \coord into a neural network effectively lifts the burden of creating high frequencies from the network, resulting in higher-quality fits of complex signals.

\paragraph{Lipschitz continuity}
A Lipschitz-continuous function is limited in how fast it can change. Formally, for this class of functions, there exists a Lipschitz bound $\lipschitzbound \geq 0$ such that
\begin{equation}
    \label{eq:lipschitz_definition}
    \|
        \signal(\coord_1) - \signal(\coord_2)
    \|_p
    \leq
    \lipschitzbound
    \|
        \coord_1 - \coord_2
    \|_p
\end{equation}
for all possible $\coord_1$ and $\coord_2$ and an arbitrary choice of $p$. Intuitively, moving a certain distance in the function's domain is guaranteed to result in a bounded change of function values.

If \signal is implemented using a fully-connected network (MLP) with \layercount layers and 1-Lipschitz activation functions (\eg ReLU), an upper Lipschitz bound is given by \cite{gouk2021regularisation}
\begin{equation}
    \label{eq:lipschitz_via_matrix}
    \lipschitzbound
    =
    \prod_{k=1}^\layercount
    \|
        \weightmatrix_k
    \|_p,
\end{equation}
where $\weightmatrix_k$ is the (trainable) weight matrix of the $k$'th network layer. Enforcing bounded weight-matrix norms effectively imposes a \emph{fixed}, \emph{global} constraint on how rapidly \signal can change.

\section{Method}
\label{sec:method}

We seek to learn a neural field $\field(\coord, \covariance)$ that captures the full anisotropic Gaussian scale space $\scalespacesignal(\coord)$ of a signal \signal, \ie a family of Gaussian-smoothed signals with arbitray, anisotropic covariance \covariance. We consider, both, coordinates \coord and covariance matrix \covariance, continuous parameters, so that the field can be queried at \emph{any} location using \emph{any} Gaussian filter. Since computing $\signal_\covariance$ via \refEq{scale_space_signal} is intractable for all but the simplest \signal, we learn $\field(\coord, \covariance)$ self-supervised, \ie we only rely on the \emph{original} signal \signal.

To achieve this goal, we make a simple but far-reaching observation: Careful dampening of high-frequency Fourier features produces a low-pass filtered signal of high quality if the neural network representing the signal is Lipschitz-bounded. Based on this observation, we develop a novel paradigm that leverages the combined properties of modulated Fourier features and Lipschitz-continuous networks~(\refSec{concept}). Our approach relies on a neural architecture with carefully designed constraints~(\refSec{architecture}), such that training can be performed using supervision from raw, unfiltered signal samples~(\refSec{training}). The emerging continuous filter parameters are uncalibrated, since we employ an efficient method that does not explicitly execute any Gaussian smoothing during training. Therefore, after training, we perform a lightweight calibration to enable precise filtering~(\refSec{calibration}).

\subsection{Self-supervised Learning of Gaussian-smoothed Neural Fields}
\label{sec:concept}

\begin{figure*}
    \includegraphics[width=0.99\linewidth]{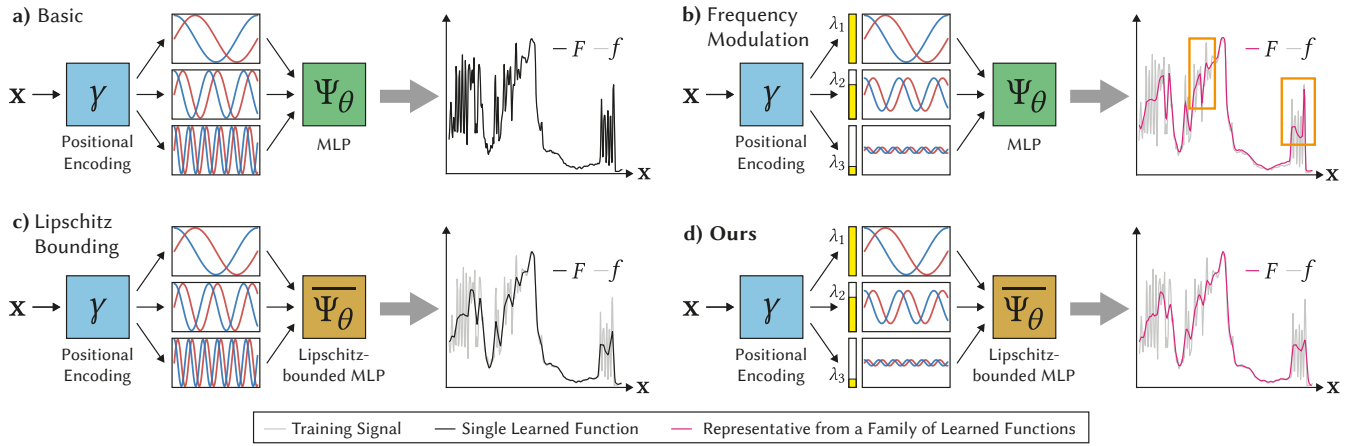}
    \caption{Four different strategies to learn a neural field \field from a signal \signal. \field takes a continuous coordinate \coord as input, which is fed into a positional encoding \positionalencoding (\refEq{positional_encoding}) that produces a set of Fourier features using cosine (blue curves) and sine (red curves) functions of different frequencies. The resulting features serve as input to an MLP \mlp that regresses \signal. (\emph{a}) The basic setup learns a faithful reconstruction of \signal (the curves for \field and \signal overlay completely), but does not allow any smoothing. (\emph{b}) Modulating the Fourier features using custom weights $\fourierfeatureweight_i$ (yellow bars) tends to remove some high frequencies, but distorts the reconstruction in an unpredictable way (orange rectangles mark incoherent spikes in \field). (\emph{c}) Employing a Lipschitz-bounded MLP \boundedmlp leads to smoothing, but it requires choosing a single fixed bound for training, lacking flexibility. (\emph{d}) Our approach combines Fourier feature modulation with Lipschitz bounding to enable controllable smoothing.}
    \label{fig:concept}
    \Description{}
\end{figure*}

We consider an established neural architecture that consists of the composition of a positional encoding using Fourier features \positionalencoding~(\refEq{positional_encoding}) with a multi-layer perceptron (MLP) \mlp:
\begin{equation}
    \label{eq:basic_network}
    \field (\coord)
    =
    \mlp \left( \positionalencoding (\coord) \right).
\end{equation}
Here, \trainableweights represents the trainable network parameters, consisting of weight matrices $\weightmatrix_k$ and bias vectors $\biasvector_k$. Based on this setup, our approach fuses two techniques that are well-known in isolation, but, to the best of our knowledge, have not yet been systematically considered in combination: First, we employ Fourier feature modulation, \ie we dampen high-frequency components of the positional encoding in \refEq{positional_encoding} using custom, frequency-dependent weights $\fourierfeatureweight_i$~\cite{barron2021mipnerf,hertz2021sape,park2021nerfies,lin2021barf}. Second, we enforce an upper Lipschitz bound of \mlp~\cite{miyato2018spectral,gouk2021regularisation,szegedy2013intriguing}. We refer to such a bounded network as \boundedmlp. To understand how this construction can help learn a controllably smooth function from a raw signal, consider four different strategies for learning a signal in \refFig{concept}.

In \refFig{concept}a, we illustrate the basic setup of \refEq{basic_network} without any modifications, \ie with $\fourierfeatureweight_i = 1$ $\forall i$ and an unbounded MLP \mlp. Unsurprisingly, we observe that training \field on \signal results in a faithful fit. Yet, we do not have any handle for creating smooth network responses here.

As a potential remedy, consider the setup in \refFig{concept}b, where we dampen the higher-frequency Fourier features of \positionalencoding. Consistent with established findings in the literature~\cite{mildenhall2020nerf,tancik2020fourfeat,muller2022instant}, we observe that fitting quality degrades. Yet, this happens in an unpredictable way, leading to incoherent high-frequency spikes in \field. This is because $\mlp$ -- depending on factors such as signal complexity and network capacity -- can compensate for missing input frequencies by forging a function with high gradients w.r.t. its inputs. Fourier feature modulation can help learn signals robustly when used progressively~\cite{hertz2021sape,lin2021barf}, or facilitate learning of a multiscale representation when supervision across scales is available~\cite{barron2021mipnerf,barron2022mip}. Yet, on its own, it is not a viable strategy for learning a smooth function from a raw supervision signal.

We now turn to an architecture with a Lipschitz-bounded MLP \boundedmlp, yet with unmodified Fourier features, depicted in \refFig{concept}c. We observe that the trained \field now indeed captures a smoother version of \signal. We seem to have achieved our goal; however, the Lipschitz bound \lipschitzbound needs to be fixed for training and is baked into the MLP. While this is a useful property for robust training~\cite{cisse2017parseval,hein2017formal} or smooth interpolation~\cite{10.1145/3528233.3530713}, this strategy does not provide any control over the smoothing once the network is trained.

\begin{figure}
	\includegraphics[width=0.999\linewidth]{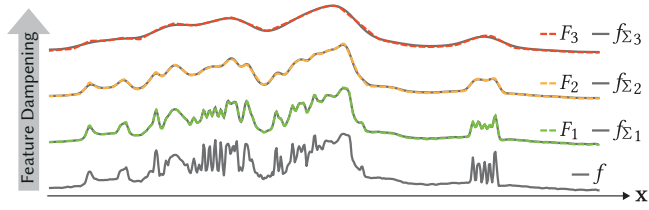}
	\caption{Given a training signal \signal (bottom row), progressive dampening of Fourier features in combination with a Lipschitz-bounded MLP allows a neural field \field learn Gaussian-smoothed versions \scalespacesignal of \signal. In the three upper rows, differently smoothed $\field_i$ (dashed colored curves) and their respective closest $\scalespacesignal_i$ (solid grey curves) are overlayed, revealing that our solution provides a faithful approximation of Gaussian filtering.}
	\label{fig:smoothing_example}
    \Description{}
\end{figure}

The above considerations motivate us to develop a new approach that combines frequency modulation with Lipschitz bounding, as shown in \refFig{concept}d. When dampening high-frequency Fourier features in this setup, the Lipschitz-bounded \boundedmlp cannot compensate for the missing frequency content, since to turn the now low-frequency encoding into a high-frequency output, it would need to produce large gradient magnitudes w.r.t. the positional encoding. Instead, \emph{it is forced to learn an \field that matches the raw \signal as closely as possible given frequency and gradient constraints.} This form of ``parameterized gradient limiting'' through modulated Fourier features facilitates controllable smoothing~(dashed colored lines \refFig{smoothing_example}).

While we intuitively expect \emph{some} form of smoothing, the exact reconstruction qualities emerging from our strategy are not obvious. However, examining reconstructions on a broad variety of real-world signals and modalities, we make a surprising, yet crucial empirical observation: \emph{The emerging smoothing is a remarkably faithful approximation of Gaussian filtering}. We extensively validate this claim in \refSec{evaluation}, but consider a rigorous theoretical justification beyond the scope of this work. In \refFig{smoothing_example}, we visually compare results \field obtained from our approach against the best-fitting Gaussian-smoothed versions \scalespacesignal of \signal (solid grey curves).

The insights developed above suggest an effective and efficient procedure for learning a Gaussian scale space in a self-supervised fashion: First, we build an architecture following \refEq{basic_network} with carefully sampled Fourier frequency vectors and a robustly Lipschitz-bounded neural network (\refSec{architecture}). Second, we train the architecture with strategically dampened Fourier features using the original signal \signal for supervision to learn an entire continuous anisotropic Gaussian scale space (\refSec{training}). Finally, we map dampening weights $\fourierfeatureweight_i$ to Gaussian covariance \covariance to enable precise filtering (\refSec{calibration}).

\subsection{Architecture}
\label{sec:architecture}

Following the reasoning developed in the previous section, we design our neural field as
\begin{equation}
    \label{eq:our_network}
    \field (\coord, \pseudocovariance)
    =
    \boundedmlp \left( \modulatedpositionalencoding (\coord, \pseudocovariance) \right),
\end{equation}
which implements two modifications to the basic setup of \refEq{basic_network}. First, we extend the positional encoding \positionalencoding to incorporate a pseudo-covariance matrix \pseudocovariance as an additional parameter, as detailed in \refSec{our_fourier_features}. Second, we use a Lipschitz-bounded MLP \boundedmlp, the construction of which is explained in \refSec{our_lipschitz}.

We emphasize that our formulation in \refEq{our_network} naturally supports spatially varying filtering, as \coord and \pseudocovariance are independent inputs to the field.

\subsubsection{Fourier Features for Filtering}
\label{sec:our_fourier_features}

\begin{figure*}
    \includegraphics[width=0.99\linewidth]{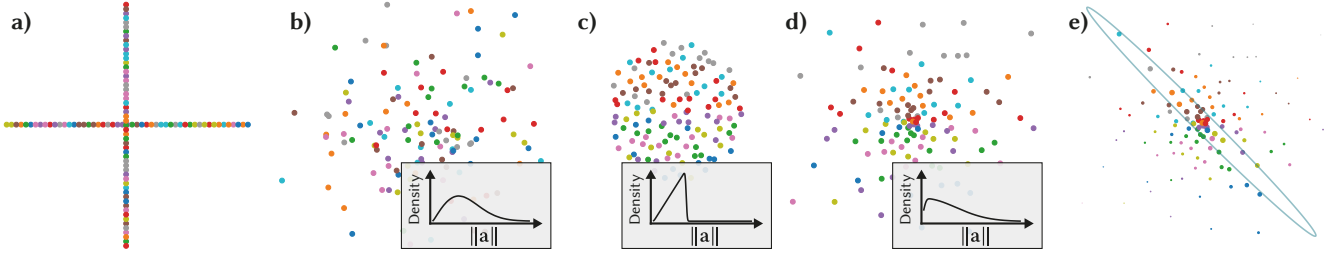}
    \caption{Distribution of Fourier frequencies \fourierfeature (colored dots). Axis-aligned frequencies (\emph{a}) cannot capture anisotropies. Uncorrelated sampling from a Gaussian distribution (\emph{b}) leads to clusters and holes, impeding filtering quality. Our approach starts with a low-discrepancy sequence (\emph{c}) and warps samples radially such that radial density follows a zero-mean Gaussian distribution (\emph{d}). The grey insets in \emph{b})-\emph{d}) show the radial density distributions of the respective point sets. In \emph{e}), we visualize an example of dampening the frequencies in \emph{d}) with a matrix \pseudocovariance (an isoline of its inverse is shown). Here, point size corresponds to dampening factors $\fourierfeatureweight_i ( \pseudocovariance )$. Our carefully distributed Fourier frequencies facilitate highly selective anisotropic filtering.}
    \label{fig:fourier_freqs}
    \Description{}
\end{figure*}

We are concerned with designing a variant of the positional encoding in \refEq{positional_encoding} that facilitates high-quality filtering through feature dampening.

We observe that the distribution of frequencies $\fourierfeature_i$ plays a crucial role in the process. Several strategies have been explored in the literature on neural field design. A popular approach relies on axis-aligned frequencies~\cite{mildenhall2020nerf,barron2021mipnerf,barron2022mip} (\refFig{fourier_freqs}a), but the lack of angular coverage does not allow arbitrary anisotropies. \citet{tancik2020fourfeat} propose to distribute frequencies following a normal distribution (\refFig{fourier_freqs}b). This leads to denser coverage, but the uncorrelated samples introduce clusters and holes. We find this uneven coverage problematic for highly selective dampening and opt for a strategy that involves stratification~\cite{niederreiter1992low}. Specifically, we use a Sobol~\shortcite{sobol1967distribution} sequence and map it to the hyperball using the method of \citet{griepentrog2008bi} (\refFig{fourier_freqs}c). We then radially warp the samples such that radially averaged sample density follows a zero-mean Gaussian distribution with variance \fourierwarpingvariance (\refFig{fourier_freqs}d). We find this shifting of sample budget towards the low frequencies a good trade-off between high-quality filtering with small-scale and large-scale kernels.

Our positional encoding needs to support anisotropically modulated Fourier features. To this end, we use a positive semi-definite pseudo-covariance matrix $\pseudocovariance \in \mathds{R}^{\indim \times \indim}$ to obtain frequency-dependent dampening factors $\fourierfeatureweight_i$ of the individual components in \refEq{positional_encoding} (\refFig{fourier_freqs}e):
\begin{equation}
    \label{eq:dampening}
    \fourierfeatureweight_i ( \pseudocovariance )
    =
    \exp \left(
        -\sqrt{
            \fourierfeature_i^T \pseudocovariance \fourierfeature_i
        }
    \right).
\end{equation}
Notice how \pseudocovariance is used without inversion here, in contrast to \covariance in \refEq{scale_space_signal}. This is a direct consequence of the reciprocal relationship of covariance in the primal and the Fourier domain~\cite{brigham1988fast}. Consider filtering a 2D signal with stronger horizontal than vertical smoothing. The convolution in the primal domain requires a kernel with higher variance in the horizontal direction, while the corresponding multiplication in the Fourier domain needs to dampen horizontal frequencies more strongly, leading to vertically elongated covariance.

Different from very similar existing techniques for supervised anti-aliasing based on axis-aligned frequencies~\cite{barron2021mipnerf,barron2022mip}, our dampening operates on Fourier frequencies that occupy the entire \indim-dimensional space, enabling arbitrary anisotropic filtering. In \refSec{calibration}, we describe how to obtain filtering results with Gaussian covariance \covariance from pseudo-covariance \pseudocovariance.

\subsubsection{Robust Lipschitz Bounding}
\label{sec:our_lipschitz}

\begin{wrapfigure}{r}{0.15\textwidth}
    \vspace{-1mm}
    \begin{center}
        \includegraphics[width=0.15\textwidth]{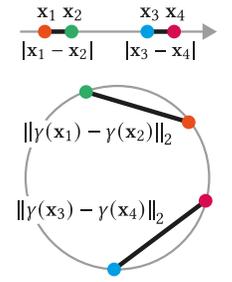}
    \end{center}
    \vspace{-2mm}
    \caption{Distances (black lines) before (top) and after (bottom) positional encoding.}
    \label{fig:why_l2}
    \Description{}
\end{wrapfigure}

We require the MLP \boundedmlp in \refEq{our_network} to be Lipschitz-bounded. \refEq{lipschitz_definition} gives us the freedom to use any $p$-norm, but we choose $p=2$, because only this choice retains spatially invariant bounding after adding the positional encoding. To understand this connection, consider a setting with a one-dimensional input coordinate \coord. Now consider a coordinate pair $(\coord_1, \coord_2)$ and a shifted version of it $(\coord_3, \coord_4)$ (\refFig{why_l2}, top). If we applied \boundedmlp directly to these coordinates, the right-hand side of \refEq{lipschitz_definition} tells us that because $\coord_1$ and $\coord_2$ have the same distance as $\coord_3$ and $\coord_4$, a fixed Lipschitz bound $c$ results in the same smoothing, \ie the bounding is spatially invariant. However, our situation is different: \boundedmlp does not operate on raw coordinates, but on their positional encoding $\positionalencoding(\coord)$, where each $\coord$ is mapped to a location on a circle (\refFig{why_l2}, bottom). Only if equidistant coordinates remain equidistant after positional encoding, formally
\begin{equation}
    \label{eq:equidistant_positional_encodings}
    \| \positionalencoding(\coord_1) - \positionalencoding(\coord_2) \|_p 
    \stackrel{!}{=} 
    \| \positionalencoding(\coord_3) - \positionalencoding(\coord_4) \|_p,
\end{equation}
the property that a specific $c$ has the same effect across the entire domain is maintained. \refEq{equidistant_positional_encodings} is only fulfilled by the isotropic 2-norm.

Following \refEq{lipschitz_via_matrix}, choosing $p=2$ translates into an MLP \boundedmlp whose weight matrices have bounded \emph{spectral} norms $\| \weightmatrix_k \|_2$. We pick a Lipschitz bound of $\lipschitzbound = 1$, which can be satisfied by individually constraining the spectral norm of each weight matrix to at most 1.

We parameterize each (arbitrarily-shaped) weight matrix using the singular value decomposition (SVD) $\weightmatrix_k = \leftsingularmatrix_k \singularmatrix_k \rightsingularmatrix_k^T$, where $\leftsingularmatrix_k$ and $\rightsingularmatrix_k$ are orthogonal matrices, and $\singularmatrix_k$ is a diagonal matrix containing the non-negative singular values of $\weightmatrix_k$. Using this decomposition, $\| \weightmatrix_k \|_2 \leq 1$ can be achieved by constraining the trainable parameters on the diagonal of $\singularmatrix_k$ using a sigmoid function. To parameterize $\leftsingularmatrix_k$ and $\rightsingularmatrix_k$, we capitalize on the fact that the matrix exponential of a skew-symmetric matrix results in an orthogonal matrix~\cite{lezcano2019cheap,hyland2017learning}. Concretely, for each $\leftsingularmatrix_k$ and $\rightsingularmatrix_k$, we arrange a suitable number of trainable parameters into skew symmetric matrices $\skewsymmetricmatrix_{\leftsingularmatrix,k} = -\skewsymmetricmatrix_{\leftsingularmatrix,k}^T$ and $\skewsymmetricmatrix_{\rightsingularmatrix,k} = -\skewsymmetricmatrix_{\rightsingularmatrix,k}^T$ and compute
\begin{equation}
    \label{eq:orthonormal_parameterization}
    \leftsingularmatrix_k = \exp(\skewsymmetricmatrix_{\leftsingularmatrix,k})
    \quad \text{and} \quad
    \rightsingularmatrix_k = \exp(\skewsymmetricmatrix_{\rightsingularmatrix,k}).
\end{equation}
Using this parameterization of weight matrices (\refFig{parameterization}), all trainable parameters of \boundedmlp can be optimized robustly and in an unconstrained fashion, while always resulting in a Lipschitz bound $\lipschitzbound \leq 1$. 

\begin{figure}[h]
    \includegraphics[width=0.99\linewidth]{figures/parameterization_01.ai}
    \caption{Our parameterization for a Lipschitz-bounded $\weightmatrix_k \in \mathds{R}^{3 \times 3}$, containing nine trainable parameters $\{ \trainableweights_1, \ldots, \trainableweights_9 \}$ that can be freely optimized.}
    \label{fig:parameterization}
    \Description{}
\end{figure}

Special care has to be taken in the case of $\outdim > 1$, \ie when \boundedmlp has multiple output channels. Recall that the definition of Lipschitz continuity in \refEq{lipschitz_definition} relies on a norm of differences between function values. One undesired way to more easily satisfy \refEq{lipschitz_definition} for $p=2$ is to lower the value of the norm on the left-hand side by producing similar function values across output channels. This is because, in expectation, Euclidean distances of point pairs on the main diagonal of $\mathds{R}^\outdim$ are shorter than distances of point pairs in full \outdim-dimensional space. In practice, we observe that, without accounting for this effect, our models tend to decrease variance between output channels, \eg they produce washed-out colors in RGB images. Fortunately, there is a simple remedy for this problem: We treat the rows of the last weight matrix $\weightmatrix_\layercount$ in \boundedmlp individually. Specifically, we replace the SVD-based parameterization of $\weightmatrix_\layercount$ by a simple row-wise $\ell_2$-normalization. This straightforward modification eliminates all undesired cross-channel contamination.

\subsection{Training}
\label{sec:training}

We train our neural Gaussian scale-space fields using the loss
\begin{equation}
    \label{eq:loss}
    \mathcal{L}
    =
    \mathbb E_{\coord, \pseudocovariance}
    \left\lVert
    \field ( \coord, \pseudocovariance )
    -
    \signal ( \coord )
    \right\lVert_2^2.
\end{equation}
It merely requires stochastically sampling coordinates \coord and pseudo-covariances \pseudocovariance to produce a filtered network output and comparing it against \emph{original} signal samples $\signal(\coord)$. We emphasize that our training does not require \scalespacesignal, thereby completely avoiding costly manual filtering per \refEq{scale_space_signal} or any approximation thereof.

As all \pseudocovariance need to be positive semi-definite, we sample them using the eigendecomposition $\pseudocovariance = \eigenvectormatrix \eigenvaluematrix \eigenvectormatrix^T$, where $\eigenvectormatrix, \eigenvaluematrix \in \mathds{R}^{\indim \times \indim}$. Specifically, we uniformly sample an orthonormal set of eigenvectors \eigenvectormatrix, and log-uniformly sample corresponding non-negative eigenvalues, arranged into the diagonal matrix \eigenvaluematrix.

The network parameters \trainableweights are optimized using Adam~\cite{kingma2014adam} with default parameters.

\subsection{Variance Calibration}
\label{sec:calibration}

Once trained, our neural field $\field(\coord, \pseudocovariance)$ in \refEq{our_network} captures a scale space, where the degree of smoothing is steered by modulation of Fourier features via pseudo-covariance \pseudocovariance. However, there is no guarantee at all that a particular choice of \pseudocovariance results in a Gaussian-filtered function with covariance $\covariance = \pseudocovariance$.

\begin{wrapfigure}{r}{0.13\textwidth}
    \vspace{-4mm}
    \begin{center}
        \includegraphics[width=0.13\textwidth]{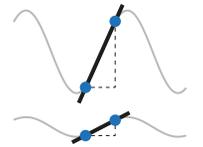}
    \end{center}
    \vspace{-2mm}
    \caption{Lipschitz bounds (see text).}
    \label{fig:why_calibration}
    \vspace{-3mm}
    \Description{}
\end{wrapfigure}

Importantly, we find that the relationship between \covariance and \pseudocovariance depends on the signal \signal itself. Consider two sine waves of the same frequency but with different amplitudes~(\refFig{why_calibration}). Gaussian filtering of these signals per \refEq{scale_space_signal} gives two identical smoothed signals up to the original difference in amplitude. Our approach does not exhibit this kind of invariance. The original low-amplitude signal has a lower Lipschitz bound (slope of the black lines in \refFig{why_calibration}) and, thus, requires more aggressive bounding to achieve the same degree of smoothing as the high-amplitude signal. Therefore, the same \pseudocovariance in \refEq{dampening} will have different effects when learning scale spaces of the two waves.

Since our ultimate goal is to produce filtering results with control over covariance that is as precise as possible, we seek to find a signal-specific calibration function $\covariancetransformation(\covariance) = \pseudocovariance$ that allows us to obtain our final Gaussian scale-space field:
\begin{equation}
    \label{eq:our_final_network}
    \field (\coord, \covariance)
    =
    \boundedmlp 
    \left( 
        \modulatedpositionalencoding (\coord, \covariancetransformation(\covariance)) 
    \right).
\end{equation}
Notice that \covariancetransformation is injected into the pipeline \emph{after} training.

\paragraph{Calibration}
We design a lightweight calibration scheme for determining \covariancetransformation that is applicable to any signal modality. First, we empirically observe that the discrepancy between \covariance and \pseudocovariance stems from a difference in \emph{isotropic} scale, while anisotropies are captured faithfully. Consequently, our calibration only considers matrices of the form $\covariance = \variance \mathbf{I}$ and $\pseudocovariance = \pseudovariance \mathbf{I}$, where $\mathbf{I}$ is the $\indim \times \indim$ identity matrix, and $\variance, \pseudovariance \in \mathds{R}_{\geq 0}$ are variance and pseudo-variance, respectively.

\begin{figure}[b]
    \includegraphics[width=0.99\linewidth]{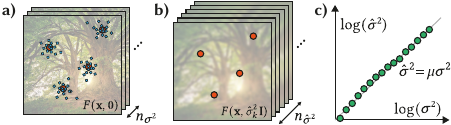}
    \caption{Our variance calibration is based on pilot coordinates $\coord_i$ (red points). (\emph{a}) We use Monte Carlo samples (blue points) to estimate ground-truth smoothing based on our field when no feature dampening is applied. (\emph{b}) We compute a sequence of differently smoothed network responses to be matched against the ground-truth from \emph{a}). (\emph{c}) The obtained variance--pseudo-variance pairs (green points) exhibit a linear relationship.}
    \label{fig:calibration}
    \Description{}
\end{figure}

We rely on computing a small number of Monte Carlo estimates of Gaussian smoothing that serve as ground truth and can be matched against our trained field. Specifically, we consider a set of $\coordsamples=64$ random pilot coordinates $\coord_i$, and a set of $\variancesamples=16$ log-uniformly spaced variances $\variance_j$. For each combination of $\coord_i$ and $\variance_j$, we compute a Monte Carlo estimate of Gaussian smoothing based on $\mcsamples=2000$ samples from $\field(\coord, \mathbf{0})$, \ie our trained field without any feature dampening (\refFig{calibration}a):
\begin{equation}
    \label{eq:mc_estimate}
    \field_{i, j}
    =
    \frac{1}{\mcsamples}
    \sum_{\offset \sim \mathcal{N}(\mathbf{0}, \variance_j\mathbf{I})}
    \field(\coord_i - \offset, \mathbf{0}).
\end{equation}
In addition, we consider a set of $\pseudovariancesamples = 256$ log-uniformly spaced pseudo-variances $\pseudovariance_k$ and compute, for each $\coord_i$ (\refFig{calibration}b),
\begin{equation}
    \label{eq:mc_candidate}
    \hat{\field}_{i, k}
    =
    \field(\coord_i, \pseudovariance_k \mathbf{I}).
\end{equation}

For each variance $\variance_j$, we now find the pseudo-variance $\pseudovariance_{k_j}$ that results in the lowest error across pilot coordinates $\coord_i$:
\begin{equation}
    \label{eq:mse}
    k_j
    =
    \argmin_k
    \sum_{i=1}^\coordsamples
    \|
        \field_{i, j} - \hat{\field}_{i, k}
    \|_2^2.
\end{equation}
Our final task is to regress the transformation \covariancetransformation that maps variances $\variance_j$ to their corresponding pseudo-variances $\pseudovariance_{k_j}$. We observe a strong linear relationship (\refFig{calibration}c), so we choose $\pseudovariance = \covariancetransformation(\variance) = \correctionfactor \variance$, where $\mu \in \mathds{R}$. Taking the logarithmic spacing of our samples into account, we regress
\begin{equation}
    \correctionfactor
    =
    \left(
        \prod_{j=1}^{\variancesamples}
        \frac{\pseudovariance_{k_j}}{\variance_j}
    \right)^{\frac{1}{\variancesamples}}.
\end{equation}
Application to the full-covariance setting gives our final calibration:
\begin{equation}
    \pseudocovariance = \covariancetransformation(\covariance) = \correctionfactor \covariance.
\end{equation}

\paragraph{Discussion}
The one-time estimations in \refEq{mc_estimate} and \refEq{mc_candidate} require a total of $\coordsamples \times ( \variancesamples \times \mcsamples + \pseudovariancesamples ) \approx 2M$ forward passes through our trained network, the total computation of which is instantaneous. Therefore, the entire calibration procedure imposes negligible cost compared to network training.

\section{Evaluation}
\label{sec:evaluation}

We demonstrate the performance of our approach on different modalities (\refSec{modalities}) and applications (\refSec{applications}), before analyzing individual components of our pipeline (\refSec{ablations}). Our source code and supplementary materials are available on our project page at \textcolor{ACMDarkBlue}{\url{https://neural-gaussian-scale-space-fields.mpi-inf.mpg.de}}.

\paragraph{Implementation Details}
All our signals are scaled to cover the unit domain $[-1, 1]^\indim$. Our positional encoding \positionalencoding uses 1024 Fourier features ($\fourierfeaturecount=512$). Networks \boundedmlp consist of four layers with 1024 features each, and are trained with a learning rate of 5e-4 (1e-4 for light stage data) until convergence. The variances for radial Fourier feature warping are $\fourierwarpingvariance = 2000$ for images, $\fourierwarpingvariance = 100$ for SDFs, $\fourierwarpingvariance = 500$ for light stage data, and $\fourierwarpingvariance = 50$ for optimization. Eigenvalues for \pseudocovariance during training are log-uniformly sampled in [$10^{-12}$, $10^{2}$]. We have implemented our method in PyTorch~\cite{paszke2017automatic}.

\paragraph{Baselines}
We quantitatively and qualitatively compare our filtering results against several baselines, while a converged Monte Carlo estimate of \refEq{scale_space_signal} serves as ground truth.

BACON~\cite{lindell2022bacon}, MINER~\cite{saragadam2022miner}, and PNF~\cite{yang2022polynomial} learn neural multiscale representations, where intermediate network outputs constitute a \emph{discrete} set of low-pass filtered versions of the original signal. Following \citet{Nsampi2023NeuralFC}, we linearly combine these intermediate outputs using coefficients that we optimize per signal to best match the filtered ground truth. Only PNF supports anisotropic filtering using a discretization of orientation. MINER requires prefiltered input during training.

We further consider INSP~\cite{xu2022signal}, which performs signal processing of a trained neural field via a dedicated filtering network. Each filter kernel requires training a separate filtering network, while our approach supports a continuous family of filter kernels.

Finally, we compare against NFC~\cite{Nsampi2023NeuralFC}, which allows filtering based on a learned integral field that needs to be queried hundreds of times per output coordinate. While this method supports continuous axis-aligned scaling of filter kernels, general anisotropic kernels require individual optimizations, leading to a discretization of kernels in the anisotropic setting. In contrast, our approach handles arbitrary anisotropic Gaussian kernels and produces filtered results using a single forward pass. We obtain best results for NFC when using piecewise linear models for 2D isotropic filtering, and piecewise constant models in all other cases.

All methods differ in their (implicit) treatment of signal boundaries. To facilitate a meaningful quantitative comparison, we crop all results such that the boundary does not influence the evaluation. Qualitative results always show uncropped signals.

Regardless of whether we evaluate isotropic or anisotropic filtering capabilities, our scale-space fields are always trained using the complete anisotropic pipeline as described in \refSec{method}.

\begin{table*}[p]
    \centering
    \caption{\textbf{Image} quality of filtering with different \textbf{isotropic} kernels (columns) for different methods (rows). ``\coord-cont.'' and ``\variance-cont.'' indicate, whether the method is continuous in the spatial and the kernel domain, respectively. Bold and underlined numbers denote the best and second-best method, respectively.}
    \renewcommand{\tabcolsep}{0.1cm}
    \label{tab:quant_img_iso}
    \begin{tabular}{lccrrrrrrrrrrrrrrrr}
        \toprule
        \multirow{2}{*}{Method} & \multirow{2}{*}{\coord-cont.} & \multirow{2}{*}{\variance-cont.} & \multicolumn{3}{c}{$\sigma^2=0$} & \multicolumn{3}{c}{$\sigma^2=10^{-4}$} & \multicolumn{3}{c}{$\sigma^2=10^{-3}$} & \multicolumn{3}{c}{$\sigma^2=10^{-2}$} & \multicolumn{3}{c}{$\sigma^2=10^{-1}$} \\
        \cmidrule(lr){4-6}
        \cmidrule(lr){7-9}
        \cmidrule(lr){10-12}
        \cmidrule(lr){13-15}
        \cmidrule(lr){16-18}
        & & & \footnotesize{PSNR$\uparrow$} & \footnotesize{LPIPS$\downarrow$} & \footnotesize{SSIM$\uparrow$} & \footnotesize{PSNR$\uparrow$} & \footnotesize{LPIPS$\downarrow$} & \footnotesize{SSIM$\uparrow$} & \footnotesize{PSNR$\uparrow$} & \footnotesize{LPIPS$\downarrow$} & \footnotesize{SSIM$\uparrow$} & \footnotesize{PSNR$\uparrow$} & \footnotesize{LPIPS$\downarrow$} & \footnotesize{SSIM$\uparrow$} & \footnotesize{PSNR$\uparrow$} & \footnotesize{LPIPS$\downarrow$} & \footnotesize{SSIM$\uparrow$} \\
        \midrule
        BACON & \checkmark & \xmark & 32.89 & 0.308 & 0.823 & \textbf{38.95} & 0.235 & \textbf{0.955} & \underline{36.48} & 0.123 & 0.953 & 30.59 & 0.086 & 0.895 & 25.36 & 0.100 & 0.601 \\
        MINER & \checkmark & \xmark & \textbf{41.19} & \textbf{0.088} & \textbf{0.963} & \underline{37.38} & 0.259 & \underline{0.945} & \textbf{36.99} & 0.097 & \textbf{0.959} & 25.89 & 0.205 & 0.815 & 24.38 & 0.156 & 0.567 \\
        INSP & \checkmark & \xmark & 30.57 & 0.454 & 0.770 & 30.14 & 0.420 & 0.838 & 23.77 & 0.546 & 0.725 & 20.75 & 0.546 & 0.627 & 23.37 & 0.381 & 0.633 \\
        NFC & \checkmark & \checkmark & 20.75 & 0.703 & 0.533 & 26.49 & \underline{0.224} & 0.839 & 36.05 & \textbf{0.071} & 0.949 & \textbf{39.74} & \textbf{0.011} & \textbf{0.965} & \textbf{41.06} & \textbf{0.006} & \textbf{0.965} \\
        \textbf{Ours} & \checkmark & \checkmark & \underline{33.85} & \underline{0.305} & \underline{0.854} & 35.05 & \textbf{0.207} & 0.942 & 34.74 & \underline{0.077} & \underline{0.954} & \underline{35.06} & \underline{0.023} & \underline{0.949} & \underline{34.99} & \underline{0.020} & \underline{0.878} \\
        \bottomrule
    \end{tabular}
\end{table*}

\begin{figure*}[p]
    \includegraphics[width=0.89\linewidth]{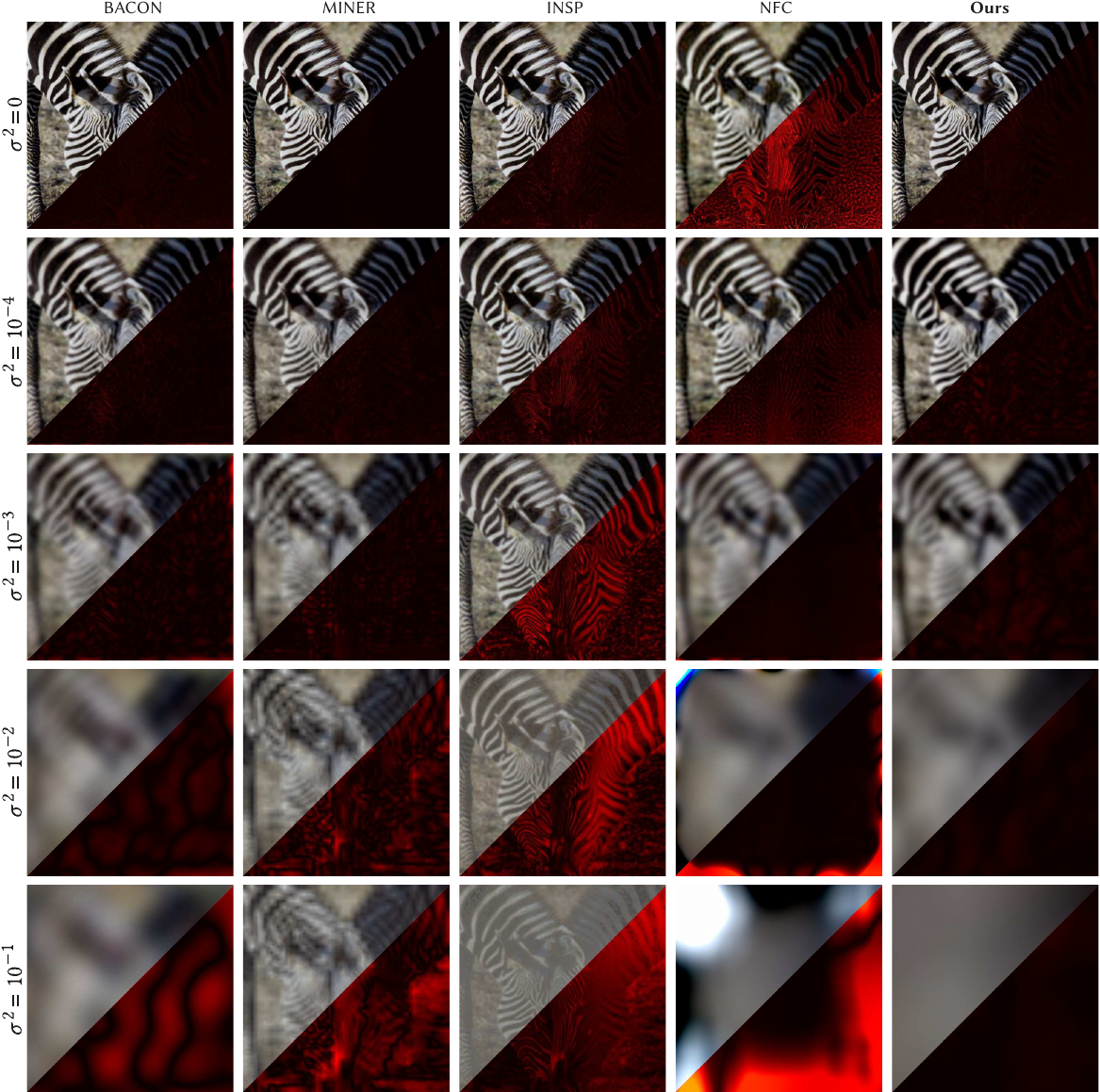 }
    \caption{Qualitative results for isotropic image filtering. We show results (upper left triangles) next to error visualizations (lower right triangles). Our supplementary materials contain more visual comparisons.}
    \label{fig:qual_img_iso}
    \Description{}
\end{figure*}

\subsection{Modalities}
\label{sec:modalities}

We extensively evaluate our method on two signal modalities relevant for visual computing: images and signed distance fields.

\subsubsection{Images}
\label{sec:images}

We consider a corpus of 100 RGB images ($\indim=2$, $\outdim=3$) at a resolution of $2048 \times 2048$ pixels, randomly selected from the Adobe FiveK dataset~\cite{fivek} and treated as continuous signals using bilinear interpolation. In a first step, we investigate \emph{isotropic} filtering on a set of five Gaussian kernels with variances $\sigma^2 \in \{0, 10^{-4}, 10^{-3}, 10^{-2}, 10^{-1}\}$, where the first configuration measures fitting quality of the original signal without any filtering. In \refTab{quant_img_iso}, we evaluate results using the image quality metrics PSNR, LPIPS~\cite{zhang2018unreasonable}, and SSIM~\cite{wang2004image}. \refFig{qual_img_iso} provides a corresponding qualitative comparison. We see that our approach is highly competitive across all filter sizes. While NFC provides the best results across all filter kernels of significant size, it introduces severe boundary artifacts, the effect of which we purposefully exclude in our numerical evaluations.

\begin{figure}
    \includegraphics[width=0.99\linewidth]{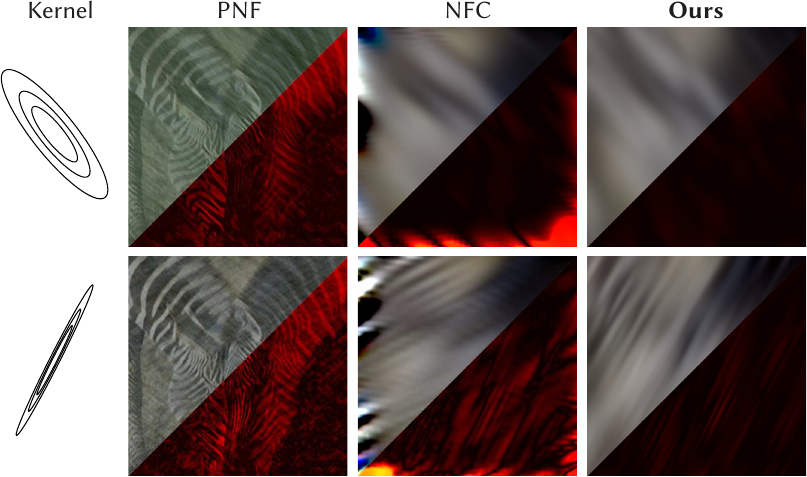}
    \caption{Qualitative results for anisotropic image filtering. Our supplementary materials contain more visual comparisons.}
    \label{fig:qual_img_aniso}
    \Description{}
\end{figure}

To evaluate performance for \emph{anisotropic} filtering, we sample 100 full covariance matrices \covariance using the scheme described in \refSec{training}, where each \covariance is evaluated on all 100 test images. We report corresponding results in \refTab{quant_img_aniso} and \refFig{qual_img_aniso} and observe that our approach outperforms all baselines on this task.

\begin{figure}
    \includegraphics[width=0.99\linewidth]{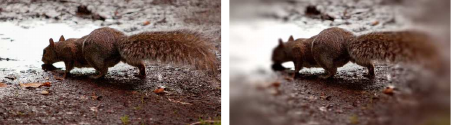}
    \caption{An original image (left) and our foveated result (right).}
    \label{fig:spatially_varying}
    \Description{}
\end{figure}

In \refFig{spatially_varying}, we show spatially varying filtering for foveated rendering. Here, the size of the filter kernel is modulated by the distance to a fixation point in the image. Our approach naturally supports such spatially varying kernels, since evaluation location \coord and filter covariance matrix \covariance are independent inputs to our model.

\begin{table}
    \centering
    \caption{\textbf{Image} quality for methods that support \textbf{anisotropic} kernels. Refer to to the caption \refTab{quant_img_iso} for details on individual columns and highlighting.}
    \label{tab:quant_img_aniso}
    \begin{tabular}{lccrrr}
        \toprule
        & \coord-cont. & \covariance-cont. & PSNR$\uparrow$ & LPIPS$\downarrow$ & SSIM$\uparrow$ \\
        \midrule
        PNF & \checkmark & \xmark & 24.15 & 0.571 & 0.704 \\
        NFC & \checkmark & \xmark & \underline{30.31} & \underline{0.094} & \underline{0.857} \\
        \textbf{Ours} & \checkmark & \checkmark & \textbf{34.82} & \textbf{0.069} & \textbf{0.940} \\
        \bottomrule
    \end{tabular}
\end{table}

\subsubsection{Signed Distance Fields}

\begin{figure*}
    \includegraphics[width=0.99\linewidth]{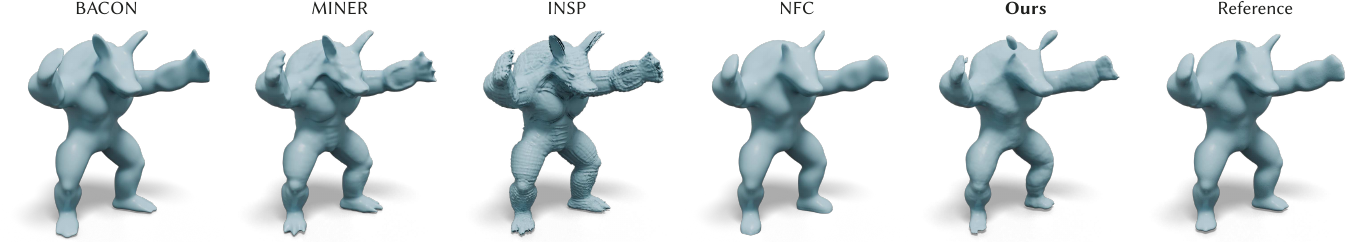}
    \caption{Qualitative results for isotropic SDF filtering on a kernel with $\sigma^2=10^{-3}$. Our supplementary materials contain more visual comparisons.}
    \label{fig:qual_geom_iso}
    \vspace{5mm}
    \Description{}
\end{figure*}

\begin{figure}
    \vspace{2mm}
    \includegraphics[width=0.99\linewidth]{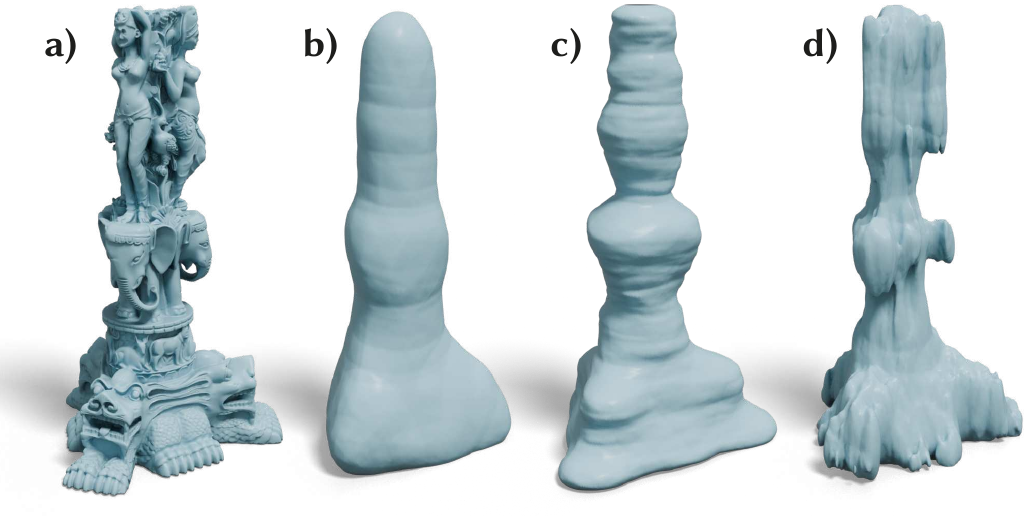}
    \caption{(\emph{a}) An original SDF. (\emph{b}) Our smoothing of \emph{a}) with an isotropic kernel, suppressing detail. (\emph{c}) Our smoothing of \emph{a}) with an anisotropic kernel, where the strength of the smoothing is weaker in the vertical direction, but strong in all other directions, allowing anisotropic structure suppression. (\emph{d}) Strong smoothing is only applied in the vertical direction.}
    \label{fig:qual_geom_aniso}
    \vspace{20mm}
    \Description{}
\end{figure}

Encoding surfaces as the zero-level-set of an SDF ($\indim=3$, $\outdim=1$) is a popular way to represent geometry~\cite{park2019deepsdf}. Our evaluation is based on four 3D models, following a similar protocol as for images. In \refTab{quant_geom_iso} and \refTab{quant_geom_aniso}, we list quantitative evaluations for isotropic and anisotropic filtering, respectively, using MSE and intersection over union (IoU) across the SDF, as well as the Chamfer distance of the reconstructed surfaces. \refFig{qual_geom_iso} and \refFig{qual_geom_aniso} show corresponding qualitative results. While results appear mostly inconclusive for the isotropic case, we outperform the only other baseline that can handle anisotropic filtering in this domain -- NFC -- by a large margin.

\begin{table*}
    \centering
    \caption{Quality of filtered \textbf{SDFs} with different \textbf{isotropic} kernels. Refer to to the caption \refTab{quant_img_iso} for details on individual columns and highlighting.}
    \label{tab:quant_geom_iso}
    \begin{tabular}{lccrrrrrrrrrrrrr}
        \toprule
        \multirow{2}{*}{Method} & \multirow{2}{*}{\coord-cont.} & \multirow{2}{*}{\variance-cont.} & \multicolumn{3}{c}{$\sigma^2=0$} & \multicolumn{3}{c}{$\sigma^2=10^{-4}$} & \multicolumn{3}{c}{$\sigma^2=10^{-3}$} & \multicolumn{3}{c}{$\sigma^2=10^{-2}$} \\
        \cmidrule(lr){4-6}
        \cmidrule(lr){7-9}
        \cmidrule(lr){10-12}
        \cmidrule(lr){13-15}
        & & & \footnotesize{MSE$\downarrow$} & \footnotesize{Cham.$\downarrow$} & \footnotesize{IoU$\uparrow$} & \footnotesize{MSE$\downarrow$} & \footnotesize{Cham.$\downarrow$} & \footnotesize{IoU$\uparrow$} & \footnotesize{MSE$\downarrow$} & \footnotesize{Cham.$\downarrow$} & \footnotesize{IoU$\uparrow$} & \footnotesize{MSE$\downarrow$} & \footnotesize{Cham.$\downarrow$} & \footnotesize{IoU$\uparrow$} \\
        \midrule
        BACON & \checkmark & \xmark & 2.5e-3 & \underline{1.3e-3} & \textbf{0.99} & 4.0e-3 & \underline{2.2e-3} & \underline{0.97} & 8.3e-2 & 1.5e-2 & 0.84 & 2.6e-4 & 4.9e-2 & 0.53 \\
        MINER & \checkmark & \xmark & \textbf{1.6e-7} & \textbf{1.1e-3} & 0.98 & \textbf{3.3e-7} & \textbf{1.4e-3} & \textbf{0.98} & \textbf{4.1e-6} & \underline{8.0e-3} & \underline{0.92} & \underline{1.8e-4} & 6.1e-2 & 0.52 \\
        INSP & \checkmark & \xmark & 1.2e-1 & 1.3e-3 & \underline{0.99} & 4.3e-2 & 4.4e-3 & 0.95 & 3.6e-2 & 1.1e-2 & 0.88 & 3.1e-2 & \underline{3.7e-2} & \underline{0.64} \\
        NFC & \checkmark & \checkmark & 3.7e-3 & 5.7e-3 & 0.89 & \underline{2.5e-5} & 4.8e-3 & 0.92 & \underline{1.4e-5} & \textbf{2.2e-3} & \textbf{0.97} & \textbf{1.0e-5} & \textbf{2.3e-2} & \textbf{0.77} \\
        \textbf{Ours} & \checkmark & \checkmark & \underline{8.3e-5} & 3.9e-3 & 0.94 & 6.0e-5 & 5.5e-3 & 0.92 & 6.5e-4 & 1.6e-2 & 0.83 & 1.1e-2 & 1.3e-1 & 0.32 \\
        \bottomrule
    \end{tabular}
\end{table*}

\begin{table}
    \centering
    \caption{\textbf{SDF} quality for methods that support \textbf{anisotropic} kernels. Refer to the caption of \refTab{quant_img_iso} for details on individual columns and highlighting.}
    \label{tab:quant_geom_aniso}
    \begin{tabular}{lccrrr}
        \toprule
        & \coord-cont. & \covariance-cont. & MSE$\downarrow$ & Cham.$\downarrow$ & IoU$\uparrow$ \\
        \midrule
        NFC & \checkmark & \xmark & 7.1e-2 & 4.6e-1 & 0.08 \\
        \textbf{Ours} & \checkmark & \checkmark & \textbf{2.8e-3} & \textbf{1.2e-1} & \textbf{0.42} \\
        \bottomrule
    \end{tabular}
\end{table}

\subsection{Applications}
\label{sec:applications}

Here, we present three applications that utilize the properties of neural Gaussian scale-space fields. First, we demonstrate anti-aliasing with texture fields, before filtering a 4D light-stage capture and showing a proof-of-concept application in the domain of continuous multiscale optimization.

\subsubsection{Texture Anti-aliasing}

Texturing a 3D mesh is a fundamental building block in many rendering and reconstruction pipelines. It requires re-sampling of a texture into screen space, which must account for spatially-varying, anisotropic minification and magnification to avoid aliasing~\cite{heckbert1986survey}. Our method enables this functionality for textures that are represented as continuous neural fields.

\begin{figure}
    \includegraphics[width=0.99\linewidth]{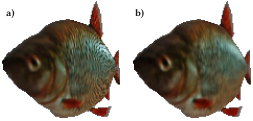}
    \caption{(\emph{a}) Rendering a textured mesh is prone to aliasing artifacts. (\emph{b}) Our method learns the continuous scale space of a texture and allows spatially varying, anisotropic pre-filtering, eliminating aliasing. Please consult our supplementary materials for video results.}
    \label{fig:texture_antialiasing}
    \Description{}
\end{figure}

In \refFig{texture_antialiasing}, we show a result using a scale-space field for texturing an object. We first learn the scale space of the texture in $uv$-coordinates, and determine the optimal anisotropic Gaussian kernel for a given camera view that results in alias-free re-sampling~\cite{heckbert1989fundamentals}. We see that our approach is successful in removing aliasing artifacts from the rendering. The supplementary materials contain a video that demonstrates view-coherent texturing.

\subsubsection{Light Stage}

A light stage allows to capture an object using different controlled illumination conditions. A structured capture, \eg using one light at a time, thus enables high-quality relighting using arbitrary environment maps in a post-process.

\begin{figure}
    \vspace{2mm}
    \includegraphics[width=0.99\linewidth]{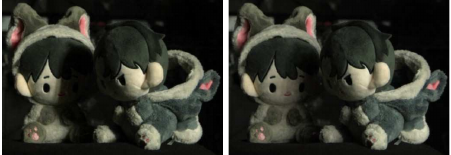}
    \caption{Filtering of 4D light stage-data leads to a smoothing of the illumination condition. Notice how the hard shadows in the original signal (left) are smoothed out in the filtered version (right).}
    \label{fig:lightstage}
    \vspace{8mm}
    \Description{}
\end{figure}

We apply our method to the 4D product space of 2D pixel coordinates and 2D spherical light directions. As demonstrated in \refFig{lightstage}, sampling a single light direction from our model at the finest scale produces hard shadows, while moving to coarser scales in light direction introduces soft shadows.

\subsubsection{Multiscale Optimization}

\begin{table*}
    \centering
    \caption{Fitting and evaluation time as well as model size on disk for different methods. ``Pref. Fit Time'' measures fitting a prefiltered image/SDF whose highest frequencies were removed by a Gaussian filter with  $\variance=10^{-4}$. ``\#Evaluations'' denotes the number of model evaluations necessary to obtain a filtered output.}
    \label{tab:performance}
    \begin{threeparttable}
        \begin{tabular}{lrrrrrrrrr}
            \toprule
            \multirow{2}{*}{Method} & \multirow{2}{*}{Disk Size} & \multicolumn{4}{c}{Image} & \multicolumn{4}{c}{SDF} \\
            \cmidrule(lr){3-6}
            \cmidrule(lr){7-10}
            & & \footnotesize{Fit Time} & \footnotesize{Pref. Fit Time} & \footnotesize{Eval. Time} & \footnotesize{\#Evaluations} & \footnotesize{Fit Time} & \footnotesize{Pref. Fit Time} & \footnotesize{Eval. Time} & \footnotesize{\#Evaluations} \\
            \midrule
            BACON & 5\,MB & 89\,s & 62\,s & 1.6\,s & 1 & 2021\,s & 1538\,s & 6.2\,s & 1 \\
            PNF & 7\,MB & 2491\,s & 1338\,s & 5.7\,s & 1 & --- & --- & --- & --- \\
            MINER & 17\,MB & 3\,s & 2\,s & 0.1\,s & 1 & 18\,s & 15\,s & 0.1\,s & 1 \\
            INSP & 4\,MB & 83\,s & 11\,s & 189.5\,s & 32\tnote{a} & 270\,s & 255\,s & 575.7\,s & 22\tnote{a} \\
            NFC & 1\,MB & ---\tnote{b} & ---\tnote{b} & 63.9\,s & 145-169 & ---\tnote{b} & ---\tnote{b} & 505.9\,s & 216-343 \\
            \midrule
            MLP & 24\,MB & 15\,s & 12\,s & 1.8\,s & 1 & 107.6\,s & 75\,s & 7.3\,s & 1 \\
            \midrule
            \textbf{Ours} & 24\,MB & 74\,s & 36\,s & 1.8\,s & 1 & 1294.1\,s & 413\,s & 7.3\,s & 1 \\
            \bottomrule
        \end{tabular}
        \begin{tablenotes}\footnotesize
            \item[a] Includes evaluations of derivative networks obtained using automatic differentiation.
            \item[b] The method did not reach the PSNR/Chamfer distance threshold.
        \end{tablenotes}
    \end{threeparttable}
\end{table*}

As a final application, we show that our scale-space fields can be used for continuous multiscale optimization. In our proof-of-concept setup, we assume that we have access to the continuous energy landscape of an optimization problem. We learn this landscape using our approach, which enables a coarse-to-fine optimization, successfully preventing the routine from getting stuck in local minima.

\begin{figure}
    \includegraphics[width=0.99\linewidth]{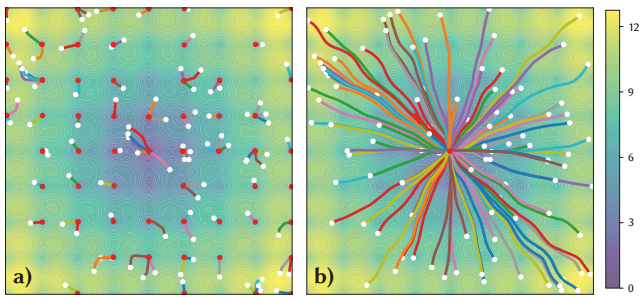}
    \caption{Our approach applied to a continuous optimization problem involving an energy landscape with multiple local minima. (\emph{a}) Performing gradient descent starting from random initializations (white dots) is prone to converging to the closest local minimum (red dots). (\emph{b}) Our scale-space field allows almost all initializations to converge to the global minimum.}
    \label{fig:optimization}
    \Description{}
\end{figure}

We demonstrate this capability using the 2D \citet{ackley} function, which consists of a global minimum surrounded by multiple local minima. A random initialization in the domain followed by gradient descent is prone to converging to one of the local minima (\refFig{optimization}a). As a remedy, we first learn the continuous scale space of the energy landscape. Then, starting from a coarse-scale landscape, we optimize the continuous location of a point using gradient descent based on automatic differentiation through our field, while progressively transitioning to finer scales. We observe that 99\% of randomly initialized points converge to the global minimum (\refFig{optimization}b), while only 5\% of points reach their goal with the single-scale baseline.

\subsection{Ablations}
\label{sec:ablations}

In this section, we analyze individual components of our method using ablational studies. We use the full anisotropic image filtering setting (\refSec{images}) for this investigation and report filtering quality for different configurations in \refTab{ablations}.

First, we are concerned with our Fourier feature dampening. We consider uncorrelated random sampling of frequencies (w/o Sobol) and removal of the frequency warping (w/o Freq. Warping). Second, we look into the Lipschitz-related components. Specifically, we consider the setup in \refFig{concept}a, where we learn a neural field and simply dampen the Fourier features after training (Freq. Scaling Only), before investigating configurations in which we plainly remove the Lipschitz bounding (w/o Lipschitz), use a looser bound (10-Lipschitz), or spectral normalization instead of our reparameterization scheme (Spectral Norm.). We also train our fields using an $\ell_1$-loss instead of using the $\ell_2$-norm in \refEq{loss} ($\ell_1$-Loss).

We observe that our full method outperforms all alternative configurations.

\begin{table}
    \centering
    \caption{Ablations.}
    \label{tab:ablations}
    \begin{tabular}{lrrr}
        \toprule
        Configuration & PSNR$\uparrow$ & LPIPS$\downarrow$ & SSIM$\uparrow$ \\
        \midrule
        w/o Sobol & 34.07 & 0.072 & 0.930 \\
        w/o Freq. Warping & 33.37 & 0.077 & 0.916 \\
        \midrule
        Freq. Scaling Only & 20.66 & 0.541 & 0.563 \\
        w/o Lipschitz & 21.37 & 0.216 & 0.634 \\
        10-Lipschitz & 32.73 & 0.076 & 0.915 \\
        Spectral Norm. & 29.08 & 0.127 & 0.868 \\
        \midrule
        $\ell_1$-Loss & 29.73 & 0.081 & 0.884 \\
        \midrule
        \textbf{Ours} & \textbf{34.82} & \textbf{0.069} & \textbf{0.940} \\
        \bottomrule
    \end{tabular}
\end{table}

\subsection{Timings and Model Size}

In \refTab{performance}, we list performance statistics across different methods. We report training times needed to achieve 30 PSNR and 0.004 Chamfer distance on unfiltered images or SDFs, respectively. We additionally measure the time and number of network evaluations required to produce a filtered output. Finally, we report model sizes when stored to disk. All experiments utilize a single NVIDIA A40 GPU.

We observe that our method is generally on par with or faster than BACON, PNF, INSP, and also NFC, which requires orders-of-magnitude more network evaluations than our approach. While MINER is faster, it is supervised on prefiltered data. A vanilla multi-layer perceptron in the form of \refEq{basic_network} (MLP) is also faster, but does not produce a scale space.

\subsection{Discussion}

\begin{figure}[b]
    \includegraphics[width=0.99\linewidth]{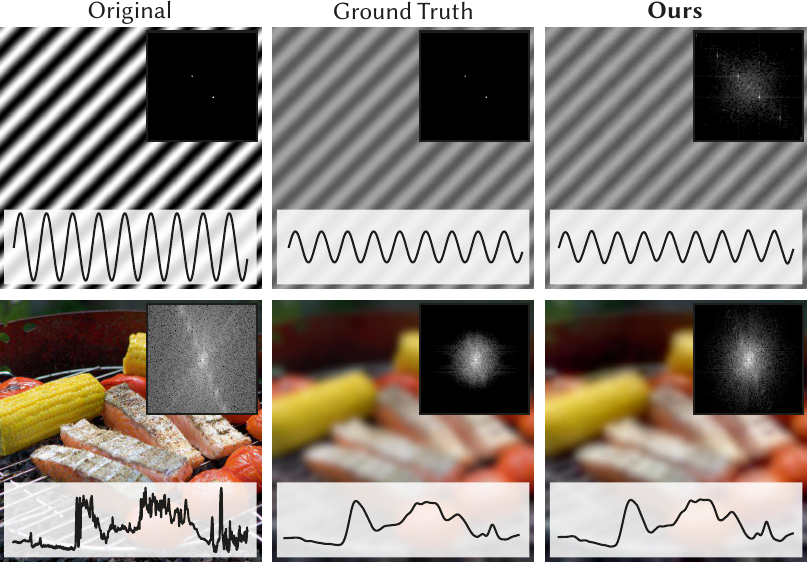}
    \caption{Original images (left) are smoothed by both a Gaussian filter (middle) and our method (right). The 1D plots show a diagonal slice through the respective image. The upper-right insets depict logarithmic spectra. While both approaches produce similar results, our method introduces slight harmonics on simple signals like the sine wave in the top row. Best viewed digitally and zoomed-in.}
    \label{fig:how_gaussian}
    \Description{}
\end{figure}

While a Gaussian filter directly dampens amplitudes of output frequencies, our method dampens amplitudes of encoding frequencies and then relies on the Lipschitz bound to carry this dampening through to the output. We demonstrate consequences of this difference in \refFig{how_gaussian}. In the top row, observe that our method reduces the amplitude of the original sine wave like a Gaussian filter. However, the peaks of our wave are sharper, approaching a sawtooth wave that one would obtain from just limiting the slope of the original wave. In the spectrum, this manifests in the emergence of harmonic frequencies. Fortunately, this effect is barely visible in more complex signals as seen in the bottom row.

Neural networks exhibit an inductive bias against learning a high-frequency output when only low Fourier encoding frequencies are present \cite{rahaman2019spectral}. The Lipschitz bound turns this bias into a hard constraint. Thus, it becomes even more important to include sufficiently high encoding frequencies, or else the unfiltered reconstruction is inadvertently bandlimited. In addition, the Lipschitz bound restricts the freedom of the neural network to learn arbitrary functions. We find increasing the network width to be an effective countermeasure.

Neural Radiance Fields \cite{mildenhall2020nerf} are a popular application of neural fields, and combining them with our method could enable cheap anti-aliasing. Unfortunately, they exhibit a very high dynamic range in volumetric density, which poses a significant challenge for a Lipschitz-bounded network to fit. Our preliminary experiments indicate that more work is needed to accommodate this specific modality.

Many of the baseline methods we consider do not generate a continuous scale space. Instead, they output a discrete set of filtered signals, which can be linearly combined to approximate all scales that lie in between. Our method similarly combines a finite set of Fourier frequencies. In contrast to these baselines, however, our combination is performed by a highly non-linear MLP, which we observe to eliminate all traces of discretization.

\section{Conclusion}
\label{sec:conclusion}

We have introduced neural Gaussian scale-space fields, a novel paradigm that allows to learn a scale space from raw data. Crucially, we have shown that a faithful approximation of a continuous, anisotropic scale space can be obtained without computing convolutions of a signal with Gaussian kernels. Our idea relies on a careful fusion of strategically dampened Fourier features in a positional encoding and a Lipschitz-bounded neural network. The approach is lightweight, efficient, and versatile, which we have demonstrated on a range of modalities and applications.

We see plenty opportunity for future work. From a theoretical perspective, it would be interesting (and ultimately necessary) to obtain a deeper understanding of \emph{why} dampened Fourier features fed into a Lipschitz-bounded network result in a good approximation of Gaussian filtering. In terms of applications, we think that our approach can potentially be a useful tool for ill-posed inverse problems in neuro-explicit frameworks, such as inverse rendering or surface reconstruction.

In light of recent efforts in continuous modeling of the real world, we hope that our neural Gaussian scale-space fields contribute a useful component to the toolbox of researchers and practitioners.

\bibliographystyle{ACM-Reference-Format}
\bibliography{paper}

\end{document}